%% file: 0_main.tex
\documentclass[sigconf]{acmart}
\AtBeginDocument{%
  \providecommand\BibTeX{{%
    \normalfont B\kern-0.5em{\scshape i\kern-0.25em b}\kern-0.8em\TeX}}}

\renewcommand\footnotetextcopyrightpermission[1]{}
\setcopyright{none}

\acmConference[]{Machine Learning on Graph (MLoG) Workshop at WSDM'22}{Feb 25th, 2022}{Virtual}
\usepackage{algorithm}
\usepackage{algorithmic}
\usepackage{xparse}
\usepackage{microtype}
\usepackage{subfig}
\usepackage{enumitem}
\usepackage{xcolor}
\usepackage{booktabs} 
\usepackage{lipsum}
\usepackage{multirow}
\usepackage[symbol]{footmisc}
\include{math_commands}

\usepackage{cleveref}

\include{my_commands}

\begin{document}

\title{Lightweight Compositional Embeddings \\for Incremental Streaming Recommendation}

\author{Mengyue Hang$^{1}$, Tobias Schnabel$^{2}$, Longqi Yang$^{2}$, Jennifer Neville$^{1,2}$} 
\affiliation{%
  \institution{$^{1}$Purdue University, $^{2}$Microsoft}
  \city{}
  \country{}
}
\email{hangm@purdue.edu, tobias.schnabel@microsoft.com, loy@microsoft.com, neville@purdue.edu}




\renewcommand{\shortauthors}{Hang and Schnabel, et al.}

\begin{abstract}
Most work in graph-based recommender systems considers a {\em static} setting where all information about test nodes (i.e., users and items) is available upfront at training time. However, this static setting makes little sense for many real world applications where data comes in continuously as a stream of new edges and  nodes, and one has to update model predictions incrementally to reflect the latest state. To fully capitalize on the newly available data in the stream, recent graph-based recommendation models would need to be repeatedly retrained, which is infeasible in practice. 
  In this paper, we study the graph-based streaming recommendation setting and
  propose a compositional  recommendation model---Lightweight Compositional Embedding (LCE)---that supports incremental updates under low computational cost.  Instead of learning explicit embeddings for the full set of nodes, LCE learns explicit embeddings for only a subset of nodes and represents the other nodes {\em implicitly}, through a composition function based on their interactions in the graph. 
  This provides an effective, yet efficient, means to leverage streaming graph data when one node type (e.g., items) is more amenable to static representation.
  We conduct an extensive empirical study to compare LCE to a set of competitive baselines on three large-scale user-item recommendation datasets with interactions
  under a streaming setting. The results demonstrate the superior performance of LCE, showing that it achieves nearly skyline performance with significantly fewer parameters than alternative graph-based models.
\end{abstract}


\maketitle

\input{1_intro}

\input{2_problem}

\input{3_method}

\input{4_exp}

\input{5_related}

\input{6_conclusion}
\bibliographystyle{ACM-Reference-Format}
\bibliography{references}

\appendix

\input{7_appendix}

\end{document}

%% file: math_commands.tex
\usepackage{amsmath,amsfonts,bm,dsfont}

\def\eqref#1{equation~\ref{#1}}

\def\1{\bm{1}}

\DeclareMathAlphabet{\mathsfit}{\encodingdefault}{\sfdefault}{m}{sl}
\SetMathAlphabet{\mathsfit}{bold}{\encodingdefault}{\sfdefault}{bx}{n}

\def\cN{{\mathcal{N}}}

\DeclareMathOperator*{\argmax}{arg\,max}

\usepackage{amsmath}
\usepackage{textcase}
\usepackage{soul}
\usepackage{enumitem}
\usepackage{indentfirst}
\usepackage{multirow}

\renewcommand{\vec}[1]{\boldsymbol{\mathrm{#1}}}

\newcommand{\uvec}{{\mathbf{z}}_u}
\newcommand{\umvec}{\mathbf{\tilde{z}}_u}
\NewDocumentCommand\wvec{o}{\mathbf{z}_w
\IfValueT{#1}{^{#1}}
}

\NewDocumentCommand\vvec{o}{\mathbf{z}_v
\IfValueT{#1}{^{#1}}
}

\NewDocumentCommand\wmat{o}{\mathbf{Z}_W
\IfValueT{#1}{^{#1}}
}
\NewDocumentCommand\umat{o}{\mathbf{Z}_U
\IfValueT{#1}{^{#1}}
}
\NewDocumentCommand\ummat{o}{\tilde{\mathbf{Z}}_U
\IfValueT{#1}{^{#1}}
}

\newcommand{\normof}[1]{\|#1\|}

\providecommand{\vz}{\ensuremath{\vec{z}}}

%% file: my_commands.tex
\def\cE{E}
\def\cV{V}

\makeatletter
\def\moverlay{\mathpalette\mov@rlay}
\def\mov@rlay#1#2{\leavevmode\vtop{%
   \baselineskip\z@skip \lineskiplimit-\maxdimen
   \ialign{\hfil$\m@th#1##$\hfil\cr#2\crcr}}}
\newcommand{\charfusion}[3][\mathord]{
    #1{\ifx#1\mathop\vphantom{#2}\fi
        \mathpalette\mov@rlay{#2\cr#3}
      }
    \ifx#1\mathop\expandafter\displaylimits\fi}
\makeatother

\newcommand{\cupdot}{\charfusion[\mathbin]{\cup}{\cdot}}

%% file: 1_intro.tex
\section{Introduction}
Real-world recommender systems face a number of important challenges in practice. First, they need to be able to model the richness of user-item and user-user interactions. Graph-based recommender systems are an excellent fit for this as they frame recommendation as a link prediction task on the user-item graph~\cite{ying2018graph, he2020lightgcn, wang2019neural}.  Another substantial challenge is the dynamic nature of the interaction data that recommendations are based on. For example, on most content platforms,  
users continuously interact with items (e.g., subscribe to a channel, visit a location) and with each other (e.g., exchange messages, co-edit a document). 

Despite the fact that graph-based data often arrives in a streaming fashion, much of the work on graph-based recommendation has focused on {\em static} settings, where all information about the test nodes (users/items) is assumed to be available for model training and assumed to be constant throughout. This assumptions limits a models ability to perform well cold-start settings and also bares high computational costs in practice since models need to be retrained regularly.
In this paper we take a step towards a more realistic recommendation setting and focus on the task of \emph{top-k} recommendation under streaming data. In top-k recommendation the goal is to recommend items that match users' long-term interests and will be consumed at some point in the future. This is different from sequential (often also called session-based or dynamic) recommendation where the task is to predict the \emph{next} item(s) that a user is going to consume~\cite{kang2018self, wang2020next, xu2020inductive, sankar2018dynamic, qiu2020gag, kumar2018learning}. 


More specifically, in this paper, we formalize the problem of graph-based recommendation in an {\em incremental} streaming setting and we propose a {\em partially compositional} model to make future recommendations efficiently. 
We assume that one node type (users or items) is more amenable to ``static'' representation than the other (e.g. items). This could mean that the activity changes more slowly so that less frequent updating is needed, or the size of the node set is small enough to relearn representations regularly. Then our model will learn a static representation for one node type (e.g, items) and learn an implicit, {\em compositional} function to calculate the representation of the other nodes (e.g, users). 

Our approach, which we call Lightweight Compositional Embeddings (LCE) allows us to:
(i) Efficiently update recommendations over time without having to regularly relearn the model,
(ii) Efficiently represent embeddings with fewer parameters compared to fully-explicit models, 
(iii) Make partially inductive recommendations (i.e., for new users or items depending on the choice above). While there are other approaches that employ compositional functions, most of them are transductive and thus are unable to support incremental updates~\cite{shi2020compositional, lian2020lightrec}. \Cref{tab: model_char} in Appendix (Sec. \ref{sec: model_comp}) compares LCE with related methods in more detail.

For empirical evaluation, we employ an incremental replay protocol to effectively assess the performance of recommendation systems in real-world streaming settings. 
Our empirical results show that LCE, with partially-implicit representations, is able to effectively utilize incremental information to substantially improve performance, with a more compact model compared to alternatives. We also evaluate LCE performance when considering cold-start items and show significant improvement over the two baselines that are able to make predictions for unseen items. Finally, we investigate LCE performance in a production recommendation setting in Microsoft Teams and show improved performance over LightGCN. 
We further include ablation experiments to show that LCE is able to (i) more effectively utilize incremental information, and (ii) approach skyline performance more quickly than alternatives. 

In summary, our contributions include:
\begin{itemize}
    \item Formalization of the setting of graph-based recommendation under incremental streaming  
    \item Development of a compositional graph-based method (LCE) that supports efficient incremental updates in a partially inductive setting
    \item Empirical results demonstrating the performance of LCE achieving significant gains over a set of competitive baselines, utilizing an incremental replay evaluation protocol to capture performance differences in a streaming setting
\end{itemize}


%% file: 2_problem.tex
\section{Problem Formulation and Notation} \label{sec: prob_form}
\begin{figure*}[ht]
\vspace{-4mm}
    \centering
    \includegraphics[width=0.8\linewidth]{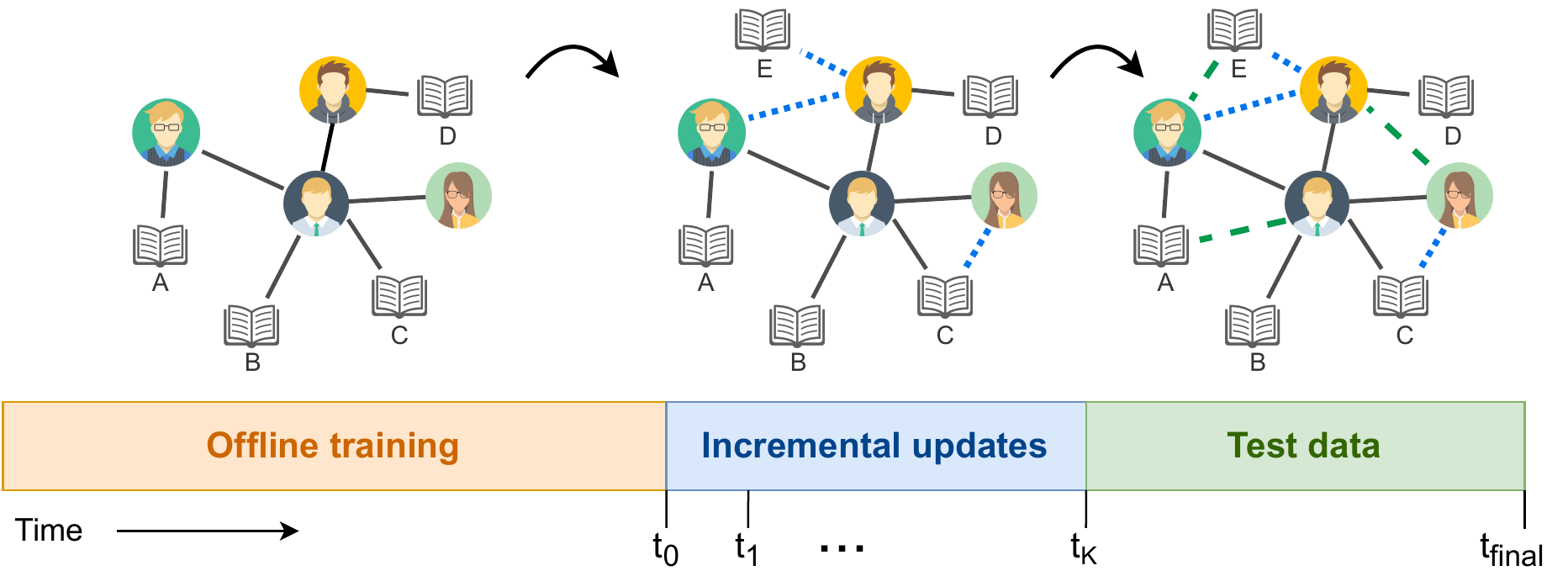}
\vspace{-4mm}
    \caption{An example graph illustrating the streaming scenario we are addressing in this paper. Given a graph at time $t_0$ --- here, between users and books --- new edges (blue) and items node (here: book E) arrive at timesteps $t_1, t_2, \ldots,$ which the model has to take into account when making predictions for edges in the final graph at $t_{final}$.}
    \label{fig:streaming_scenario}
\vspace{-3mm}
\end{figure*}


In this paper, we develop and evaluate our method in what we call the \emph{incremental streaming setting}. In it, we assume that there exists some fixed future graph $G_{final} = (\cV_{final}, \cE_{final})$ whose edges form the target of our prediction. Nodes are split into two types (users and items), $\cV = U \cupdot W$ with no edges between items as item-item interactions are rare in practice. 
We do not get to observe the entire set of edges $\cE_{final}$ or nodes $\cV_{final}$ at the start but only subsets of them $\cE_{t_0}$ (or $\cV_{t_0}$) at the beginning -- at a point in time we denote with $t_0$. Now, as time moves forward, we get to see an increasing number of edges $\cE_{t_0} \subseteq \ldots \cE_{t_K}$ and nodes $\cV_{t_0} \subseteq \ldots \cV_{t_K}$ from $G_{final}$. This process can also be seen as sampling without replacement from the edge set and nodes sets of $G_{final}$. We assume a partially inductive setting, where new nodes of only a certain type (i.e. user or item) can arrive over time, and the other type of nodes are known at $t_0$, i.e., $\cV_{t_i} = W_{t_i} \cupdot U_{final}$. 

We assume that we are allowed to train a model once for $t_0$, a step we will refer to as \textbf{offline batch training}. After that, the model goes through $K$ rounds of \textbf{online incremental updates} where it observes new edges and nodes as it is asked to produce updated predictions. More specifically, the prediction target is the \textbf{edge set} $E_{t_{final}} - E_{t_K}$ and the goal is to condition predictions on the graph $G_{t_K}$, even though we can only estimate  model parameters on $G_{t_0}$. 

\Cref{fig:streaming_scenario} illustrates the incremental streaming scenario with an evolving user-book graph where user-user edges represent trust relations and user-book edges indicate read books. The middle graph corresponds to an intermediate update step $t_i$ where a new book E and three blue dotted edges were added to the graph. The rightmost graph represents the final graph $G_{final}$ where the green edges were added and also form the prediction target.

%% file: 3_method.tex
\section{Lightweight Compositional Embeddings}
The key idea of our approach, Lightweight Compositional Embeddings (LCE), is to only explicitly learn embeddings for one set of nodes -- either $U$ or $W$ -- and then compute representations for the other set of nodes via a composition function. In practice, one would typically choose to represent the smaller node set (either $U$ or $W$) explicitly. This enables LCE to efficiently update under a streaming setting, as well as substantially reduces overall model size. During the offline initialization step, we fit the explicit embeddings and aggregation parameters in the LCE model as usual on the training set, propagating information from embeddings via simplified GCN layers. Afterwards, during each incremental update step, we re-compute the compositional embeddings with the new edges from each step and also compute new prediction scores.

\vspace{-2mm}
\subsection{LCE model}
We start with a general description of the LCE model before turning to training and inference details.
For ease of exposition, let us assume that we chose to learn explicit embeddings for $u \in U$ and represent $w \in W$ implicitly through a composition function.  LCE then represents each user $u$ via a pair of explicit embeddings $(\uvec, \umvec) \in \mathbb{R}^d \times  \mathbb{R}^d$, enabling the model to use different embeddings for aggregation and scoring. Large letters denote matrices, e.g., $\umat$ is the matrix one obtains by concatenating all $\uvec$ from $U$. Let us use $\wvec$ to denote the compositional embedding of $w \in W$ that are derived from $\umat$ via the graph interactions. 
Given representations $(\uvec, \umvec)$ and  $\wvec$, we model the probability of an edge existing between user $u$ and item $w$ as

\begin{equation} \label{eq: reco_score}
            S(u, w)  = \sigma(\umvec^T\wvec),
\end{equation}

\noindent where $\sigma$ is the well-known sigmoid function. 

{\em Inference.}
To generate recommendations we compute the scores $S(u, w)$ for user $u$ on all items $w$, and sort them in descending order to produce a ranking $w_{(1)}, w_{(2)}, \ldots,  w_{(|W|)}.$


{\em Compositional embeddings.}
To generate embeddings $\wvec$ for each $w \in W$, LCE follows the process in Eq.\ref{eq: emb}. 
First, we compose the initial embeddings for each item $w \in W$ by aggregating embeddings from the incoming edges of $w$ within a given graph $G$:
\begin{equation}
    \wvec^{(0)} = f(\{ \uvec \}_{u\in \cN_U(w;\,G)}) \label{eq: init}
\end{equation}
 Here, $\cN_U(w; G)$ is the set of neighboring users for item $w$, and $f(\cdot)$ is a composition function (e.g. mean pooling, sum pooling) which generates the input layer embedding $\wvec^{(0)}$ as the average or sum of the explicit embeddings $\uvec$ according to past user-item interactions. The impact of this choice is generally small as our empirical result shows, but we note that this is an additional hyperparameter that can be optimized.
 
 $G$ denotes the adjacency matrix of a graph which will be specified in the following two subsections.
We stack $L$ graph convolution layers to pass around information from the neighborhood. For each node $v\in U \cupdot W$, we compose its embedding $\vvec$ as follows:
\begin{equation}\label{eq: emb}
\begin{split}
    \vvec^{(1)} &= agg(\{\vz_{v'}^{(0)} \}_{v'\in \cN_{G}(v;\: G)}) \qquad \ldots \qquad \\ \vvec^{(L)} &= agg(\{\vz_{v'}^{(L-1)} \}_{v' \in \cN_{G}(v;\: G)}
\end{split}
 \end{equation}
 \begin{equation} \label{eq: item_avg}
     \wvec = avg(\wvec^{(0)}, \wvec^{(1)}, \cdots, \wvec^{(L)})
 \end{equation}
   where $\cN_{G}(v;\: G)$ is the set of all the neighbors for node $v$ in graph G that might include both users and items. The $agg$ function denotes an aggregation function corresponding to a graph convolution operator which generates $l$-th layer’s node representation from $(l-1)$-th layer's representation of the target node and its neighbor nodes. 
   
   Note that the convolution function (Eq.\ref{eq: emb}) is applied for all nodes, but the final composition (Eq.\ref{eq: item_avg}) is only used to calculate implicit embeddings for nodes $W$. 
  Similar to LightGCN~\cite{he2020lightgcn} which showed that typical aggregation functions can be suboptimal in recommendation tasks, we adopt a simplified design for the convolution layers where the aggregation function is mean pooling and the layer-wise transformation function is simply the identity mapping. 



\subsection{Offline batch training}
As mentioned before, we train the model once on a batch of offline data before it enters the streaming phase. Let $\Theta$ denote the set of all parameters in our model. 
We initialize the matrices of explicit embeddings $(\umat, \ummat)$ with random entries and then update later during training them via back propagation. In Eq.\ref{eq: emb}, we use the initial snapshot $G_{t_0}$ for the graph $G$. We learn all parameters of our model via an auxiliary prediction task where we sample a few target edges from the training edges $\cE_{train}$ in each epoch which we will try to reconstruct from the remaining input edges. We use the Bayesian Personalized Ranking (BPR) loss to encourage the relevant items to be ranked higher than the other items, as shown in Eq.\ref{eq: bpr_loss}. This is equivalent to maximizing the likelihood of existing user-item edges with negative sampling,  

    \vspace{-3mm}

    \begin{align} \label{eq: bpr_loss}
        \hat{\Theta} = \argmax_{\Theta} \sum_{u\in U} \sum_{\substack{w\in \\ \cN_W(u;\: G)}} \sum_{\substack{w' \notin \\ \cN_W(u;\: G)}} &\ln \sigma( \umvec^T\wvec - \umvec^T\vz_{w'})\nonumber \\ &+ \lambda \normof{\Theta}^2 
    \end{align}
where $\cN_W(u;\: G)$ represents the set of neighbors with type $W$ in the training graph G, which are essentially items that the user $u$ has interacted with in the past.
We use a separate validation set for hyperparameter search and refer to the appendix for details of the training procedure. 
Note model parameters $\Theta = \{\tilde{z}_u, z_u\}_{u\in U}$ are optimized with Eq.~\ref{eq: bpr_loss} and Eq.~\ref{eq: init}-\ref{eq: item_avg} show how to compute $z_w$ from $z_u$.


\subsection{Online incremental (streaming) updates} 
After offline training, we fix the explicit embeddings  $(\umat, \ummat)$ and re-compute embeddings $\wmat$ as follows. Given a new graph $G_{t_k}, k > 0$, we use the new adjacency matrix $G_t$ in place of $G$ to initialize item embeddings via Eq.\ref{eq: init} and follow Eq.\ref{eq: emb} to obtain the final embedding for $w$, again plugging in $G_{t_k}$ for $G$. Finally, we obtain new recommendation scores via Eq.\ref{eq: reco_score} on missing edges between users and items. 



%% file: 4_exp.tex
\vspace{-2mm}
\section{Empirical evaluation}

Through our experiments, we seek to answer the following research questions:
\begin{enumerate}
    \item How does our proposed method fare against a competitive set of baselines? Particularly, can the models effectively use incremental updates to improve model performance? ($\rightarrow$ Sec.~\ref{sec:expts-base})
    \begin{itemize}
        \item How much does performance improve over the offline setting? ($\rightarrow$ Tables \ref{tab: results_streaming}, \ref{tab: results_offline}, Fig. \ref{fig:online})
        \item How close do they get to skyline performance (i.e. retraining models with all available data)? ($\rightarrow$ Fig. \ref{fig:online})
        \item How do they perform when including cold-start items during streaming? ($\rightarrow$ Fig. \ref{fig:cold_start}) 
    \end{itemize} 
    \item How important are the various components of LCE? Does performance suffer if the components are removed/varied? ($\rightarrow$ Sec.~\ref{sec:expts-ablation}) How does the choice of compositional embedding affect  performance? ($\rightarrow$ Sec.~\ref{sec:explitit_vs_implicit})
    \item How will LCE perform in a production recommendation system? ($\rightarrow$ Sec.~\ref{sec:expts-prod})
\end{enumerate}

\subsection{Replay protocol}
For the overall experimental setup, we implemented the scenario introduced in \Cref{sec: prob_form} as follows. There are three larger splits of the data -- an offline dataset to train ($G_{t_0}^{train}$) on with a validation set ($G_{t_0}^{val}$), $K$ different chunks of streaming data ($G_{t_i}$), and the test data ($G_{final}$). This is also shown in \Cref{fig:streaming_scenario}.

For each method to evaluate, we first train a model $M_{t_0}$ using the training portion offline data $G_{t_0}$, and pick the best hyperparameters on the validation test $G_{t_0}^{val}$. During the streaming phase, we regard the model parameters as fixed and only feed new inputs $G_{t_i}, i = 1, \ldots, K$ to the model $M_{t_0}$. We will also report skyline performance later which is the performance of a model that was completely retrained with all data up to and including $G_{t_i}$. Let $M_{t_i}$ denote the skyline model one gets from the latter. To measure recommendation performance, we rank all possible user-item edges by their scores and measure quality with respect to all new edges in the final graph, $G_{final} \setminus G_{t_{K}}$. We report test set performance as recall@$N$ and nDCG@$N$, with both being common metrics in top-$N$ recommendation.

\subsection{Data}

We consider the following three datasets for our experiments whose statistics are shown in~\Cref{tab:dataset_stats}. They differ in the time span and user/item ratio, and the user-item graph density. 
For example, the Yelp dataset contains the reviews from 20K users for 37K items over the time period of three years, whereas the Epinions dataset contains much sparser user-item interactions with fewer users ($\sim$10K) but more items ($\sim$88K) over ten years. 
More details about each dataset are listed below:

\begin{table}[]
    \centering
    \begin{tabular}{lccc} \toprule
        Dataset & Yelp & Epinions & LibraryThing \\ \midrule
    \# users & 20,458 & 10,277& 16,894\\
    \# items & 37,552 & 87,791 & 59,079 \\
    graph density & 0.079\% & 0.023\% & 0.037\% \\
    \midrule
    offline window & 24 months & 48 months & 50 months\\
    streaming window & 3 months & 12 months & 12 months\\
    test window & 6 months & 5 years & 3 years\\ \bottomrule
        
    \end{tabular}
    \caption{Density, user-item ratios, and temporal window sizes (used for creating data splits)  for the three datasets we used during evaluation.}
    \label{tab:dataset_stats}
\end{table}

\begin{table}[tbp!]
    \centering
    \vspace{-5mm}
    \begin{tabular}{ccccc}\toprule
    method  & architecture & embeddings & model size \\ \midrule
    $RP^3_{\beta}$  & RW  & & none\\
    ALS     &  MF & user + item &   $(|U| + |W|) \times d $\\
    SLIM     &  MF & item & $|W|\times |W|$\\
    ENSFM & Deep MF  & user + item & $(|U| + |W|) \times d $\\
    LightGCN & GNN  & user + item & $(|U| + |W|) \times d $\\
    LCE & GNN  & user \emph{or} item& $2|U| \cdot d $\\ \bottomrule
    \end{tabular}
    
    \caption{Properties of baselines used in our experiment. We compare to a mix of classic factorization-based approaches and deep/graph-based models. RW=``random walk''.}
    \label{tab:method_char}
\end{table}


\begin{description}
\item[Yelp.] The Yelp dataset is adopted from the latest Yelp challenge\footnote{https://www.yelp.com/dataset} which includes time-stamped reviews from Yelp users from the year of 2017 to 2019. The local businesses like restaurants and bars are viewed as the items, and the user-item interactions are reviews given to the restaurants. 

\item[Epinions.] This dataset contains timestamped product reviews and the trust network amongst users~ \cite{tang2012mtrust}\footnote{https://www.cse.msu.edu/~tangjili/trust.html}. 

\item[LibraryThing.] Dataset including book ratings and social relationships between users, from Aug. 2005 to Aug. 2013~\cite{cai2017spmc}\footnote{https://cseweb.ucsd.edu/~jmcauley/datasets.html}. 
\end{description}

As the number of items are much more than number of users in the three datasets, we learn {\em explicit} embeddings for users and {\em compositional} embeddings for items to have a more compact model. \footnote{See \Cref{sec:explitit_vs_implicit} for an empirical assessment of the explicit v.s. implicit settings.} Also, since some baseline models (e.g. LightGCN, ENSFM) do not support cold-start items, we primarily consider a transductive setting where all items and users are known during offline training, and also introduce a partially inductive setting, which includes cold-start items, for evaluation in Sec. \ref{sec: cold_start_eval}.

\subsubsection{Data splits.} To create the splits mentioned in \Cref{fig:streaming_scenario}, we split the user-item interactions by their timestamps into three sets (i.e. offline, streaming and test) and use all the user-user edges for offline training. 
The last $10\%$ of the offline data makes up the validation set used for hyper-parameter tuning. 
The overall proportions were chosen so as to ensure that (i) we have enough users from $G_{final}$ that also appear in $G_{t_{0}}$ and (ii) have enough observations to perform the incremental streaming updates. The streaming portion was divided into three equally sets $G_{t_{1}}, G_{t_{2}}$, and $G_{t_{3}}$, and we assess model performance under increasing amounts of streaming data in~\Cref{sec: use_streaming_data}. The specific split sizes are shown in the bottom part of \Cref{tab:dataset_stats}.

\subsection{Baselines}

We consider both popular recommendation methods (ALS, SLIM, ENSFM, etc.) and graph-based method (LightGCN). \Cref{tab:method_char} lists the characteristics of the various baselines and our model. These methods differ in structure (i.e. shallow v.s. deep or graph-based),  model parameterization (i.e. whether explicitly learn user/item embeddings) resulting in the difference in model parameters. 

The implementation details (e.g. hyper-parameter tuning) of each baseline method can be found in the appendix. For fair comparison, we feed matrix-factorization based algorithms a user-(item $\cup$ user) matrix by concatenating the user-user interaction matrix with the user-item matrix.
\begin{itemize}[leftmargin=*]
    \item Top-Popu: recommend most popular items.
    \item Alternative Least Squares (ALS) \cite{hu2008collaborative, takacs2011applications}.
    \item Efficient Non-sampling Factorization Machines (ENSFM) \cite{chen2020efficient}. We represent the user-item graph as one-hot feature vectors.
    \item $RP^3_{\beta}$ \cite{christoffel2015blockbusters}: a graph vertex ranking recommendation method that re-ranks items based on 3-hop random walk transition probabilities.
    \item Sparse Linear Method (SLIM) \cite{ning2011slim}. SLIM is one of the most competitive baselines in top-$N$ recommendation.
    \item LightGCN \cite{he2020lightgcn}: LightGCN is a state-of-the-art graph-based model for collaborative filtering. To leverage the social network information, we feed the heterogeneous graph including both user-item and user-user edges instead of the bipartite graph considered in the original paper. 
    \item LCE variants: In our ablation study (see \Cref{sec:expts-ablation}), we compare to two variants of our model: "LCE-1 emb" and "LCE-1 layer". "LCE-1 emb" does not have a separate user embedding for scoring user-item pairs, but uses the user embedding generated by GCN layers instead (i.e., setting $\uvec$ with \Cref{eq: item_avg}). 
    "LCE-1 layer" uses a single GCN layer instead of three layers. 
\end{itemize}

\begin{table*}[tpb]
\centering
\resizebox{0.8\textwidth}{!}{
\begin{tabular}{lllllll} \toprule
\cmidrule(lr){2-5}
& \multicolumn{2}{c}{Yelp} &\multicolumn{2}{c}{LibraryThing}
&\multicolumn{2}{c}{Epinions}\\
\cmidrule(lr){2-3} \cmidrule(lr){4-5} \cmidrule(lr){6-7}
&  Recall@20 & nDCG@20 & Recall@20 & nDCG@20 & Recall@20 & nDCG@20 \\ \cmidrule(lr){1-1} \cmidrule(lr){2-2}\cmidrule(lr){3-3} \cmidrule(lr){4-4}\cmidrule(lr){5-5}
\cmidrule(lr){6-6}\cmidrule(lr){7-7}
Top-popu & 0.0040 & 0.0022  & 0.0009 & 0.0014 & 0.0012 & 0.0007\\
$RP^3_{\beta}$ & 0.0336 & 0.0199 & 0.0049 & 0.004 & 0.0048 & 0.0032\\
ALS                   & 0.0438 & 0.0263 & 0.0215 & 0.0202 & 0.0112 & 0.0105\\
SLIM                  & 0.0481 & 0.029 & 0.028 & 0.0269 & 0.0149 & 0.0134\\
ENSFM                 &0.0565 & 0.0344 & 0.0265 & 0.0259 & 0.0091 & 0.0079\\
LightGCN    & 0.0565 & 0.0334 & 0.024 & 0.0209 & 0.0152 & 0.0100\\
\midrule
$LCE_{mean}$ & \textbf{0.0651}$^{\star}$ & \textbf{0.0387}$^{\star}$ & 0.0299 & 0.0282 & 0.0167 & 0.0144\\
$LCE_{sum}$ & 0.0620 & 0.0361 & \textbf{0.0301} & \textbf{0.0283} & \textbf{0.0178}$^{\star}$ & \textbf{0.0148}\\
LCE-1 emb   & 0.0636 & 0.0380 & 0.0286 & 0.0242 & 0.0153 & 0.0106\\
LCE-1 layer & 0.0595 & 0.0349 & 0.0216 & 0.0198 & 0.0095 & 0.0067\\ \midrule
\% improvement & +15.22\% & +12.50\% & +7.50\% & +5.20\% & +17.11\% & +10.45\%\\
 \bottomrule
\end{tabular}
}
\caption{The streaming performance on three datasets. Our model (LCE) consistently out-performs all the other methods. LightGCN and SLIM are the two strongest baselines. Numbers with $\star$ represent significant improvement in a paired t-test at the $p < 0.05$ level compared with the best baseline.}
\label{tab: results_streaming}
\vspace{-4mm}
\end{table*}

To test each of these methods, we adopt the same temporal data split and the incremental replay evaluation protocol. The only exception is for ENSFM \cite{chen2020efficient} method. Since it does not support incremental updates, we retrain the model from scratch using both offline and streaming data for the streaming setting.

\vspace{-2mm}
\subsection{Comparison with baseline methods} \label{sec:expts-base}

\Cref{tab: results_streaming} shows the streaming recommendation performance in terms of recall@20 and nDCG@20 of baseline methods and our method (including its variants) on the three datasets. The offline performance refers to the scenario when the offline portion of our dataset, $G_{t_{0}}$ is used for training, while the streaming performance refers to when the streaming data is used for inference using the same trained model. In the streaming scenario, we report performance after seeing the last chunk of streaming data $G_{t_{3}}$. Comparing the offline and streaming results, almost all the methods are able to utilize the streaming data effectively to improve streaming performance, with the only exception of LightGCN on Yelp dataset.

Our proposed method LCE consistently out-performs all the baselines, especially when streaming data is available. To test the statistical significance of the performance gain, we conduct paired t-tests with alternative hypothesis that LCE performs better than best baseline, and mark \Cref{tab: results_streaming} with $\star$ when $p < 0.05$.

Among the baselines, LightGCN and SLIM are the two strongest models. More specifically, SLIM achieved better nDCG scores than LightGCN but lower recall scores, meaning that the sparse model (i.e. SLIM) is not as effective at retrieving all relevant items as it is for ranking them accurately. 
Note however that, under the same embedding dimension, LCE has fewer parameters than LightGCN when the size of the implicitly embedded node set is larger than that of the explicitly embedded set (see \Cref{tab:method_char}). Moreover, LCE performance is significantly better than LightGCN across the three datasets. See Appendix \Cref{sec: capacity} for more results on model capacity v.s. performance.

The ENSFM model achieved better performance than SLIM on Yelp, but not on the other two datasets. Note that we do not have additional features (e.g., user features) as input which is typical available for factorization machine-based methods, thus it might limit the strength of ENSFM in our setting. The shallow models ALS and the ``lazy'' approach $RP^3_{\beta}$ and top-popu in general perform worse than the other deep(er) models, which also indicates that more complex model (i.e. deep, graph-based) can better capture the signals from past interactions for recommendation.

\vspace{-2mm}


\subsubsection{Utilizing streaming data with incremental updates} \label{sec: use_streaming_data}
To further investigate if LCE is able to effectively utilize the streaming data comparing to the skyline performance when retraining with full data, we compare the streaming performance using incremental updates (the default setting) vs. skyline performance when retraining the model from scratch with full (i.e. offline + streaming) data.

\begin{figure*}[h!]
\centering
 \vspace{-5.mm}
\subfloat[Yelp]{
    \includegraphics[width = 5.8cm]{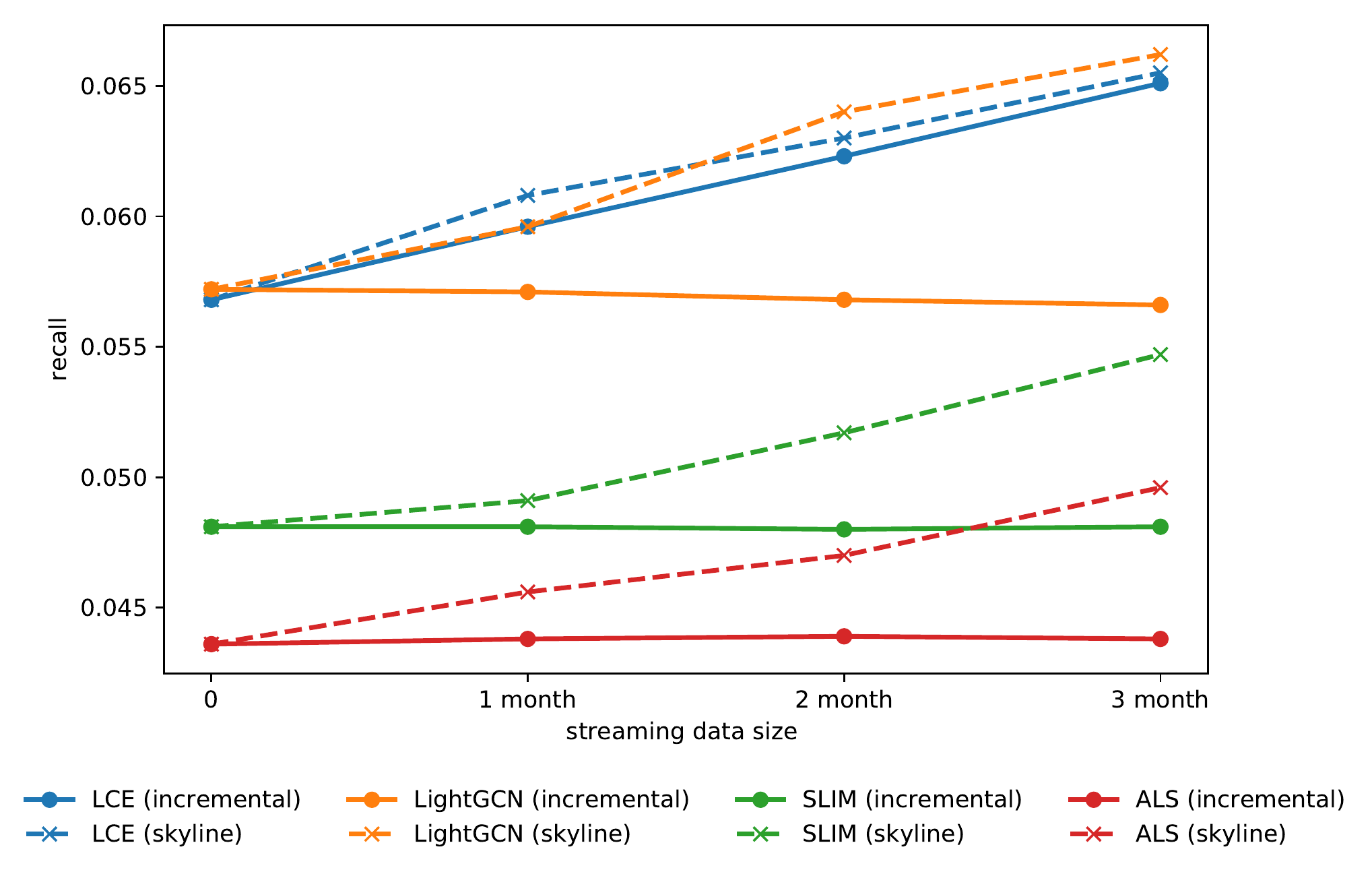}
    \label{fig:online_yelp}
}
\subfloat[LibraryThing]{
    \includegraphics[width = 5.8cm]{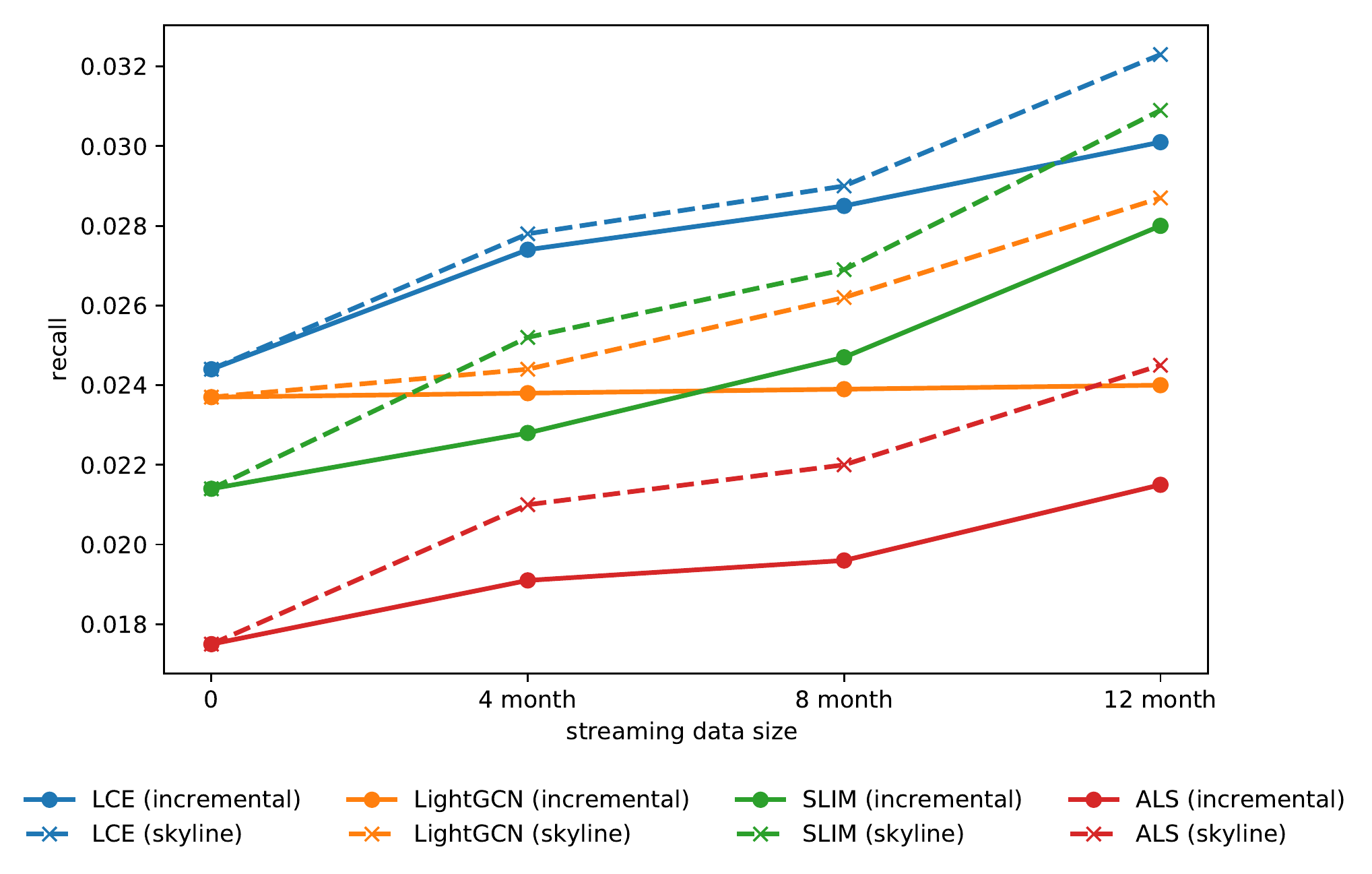}
    \label{fig:online_lthing}
} 
\subfloat[Epinions]{
    \includegraphics[width = 5.8cm]{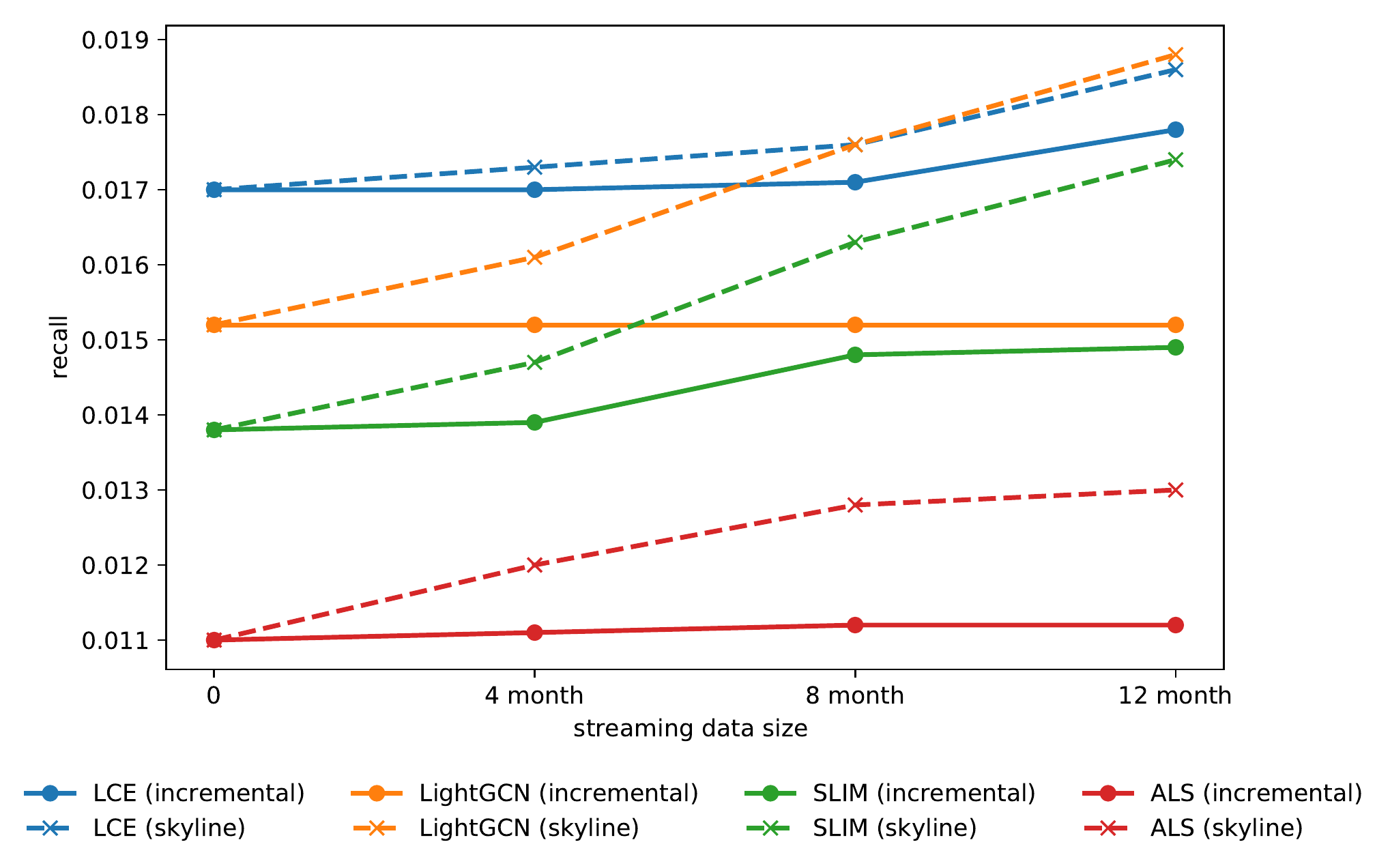}
    \label{fig:online_epin}
}
\vspace{-3.mm}
\caption[]{Incremental (solid lines) and skyline (dotted lines) performance of various methods on Yelp (\ref{fig:online_yelp}), LibraryThing (\ref{fig:online_lthing}) and Epinions (\ref{fig:online_epin}) datasets.  LCE refers to one of $LCE_{mean}$ or $LCE_{sum}$ variants with better validation performance ($LCE_{sum}$ for both datasets). On Yelp and Librarything datset, our model LCE is comparable or slightly better than the best baseline (i.e. LightGCN) when no streaming data is available, but gain becomes larger as more data comes in. The smallest gap between the two lines of our model reveals its ability to utilize streaming data via incremental updates. On Epinions dataset, our model is consistently better than the best baseline (i.e. LightGCN).}
\label{fig:online}
\end{figure*}

We show the model performance with incremental updates (solid lines) and with model retraining (dotted lines) on the three datasets in \Cref{fig:online}. Again, we can see that our method LCE consistently out-performs the three baselines (LightGCN, SLIM, and ALS), and the gains become larger as more streaming data is available for incremental updates. Furthermore, comparing the gaps between the two lines for each method, LCE has the smallest gap compared to other baselines. For example, on Yelp dataset the skyline performance of LCE and LightGCN is very close, but LCE has much better incremental performance, which means LCE can better utilize the streaming data even without retraining the model. 
Note that incremental updates are much more efficient than model retraining. Each incremental step takes a few seconds for LCE while model re-training takes a few hours.

\vspace{-1mm}
\subsubsection{Evaluating with cold-start items} \label{sec: cold_start_eval}
\begin{figure}[h!]
\centering
\vspace{-3.mm}
\subfloat[LibraryThing]{
    \includegraphics[width = 5.5cm]{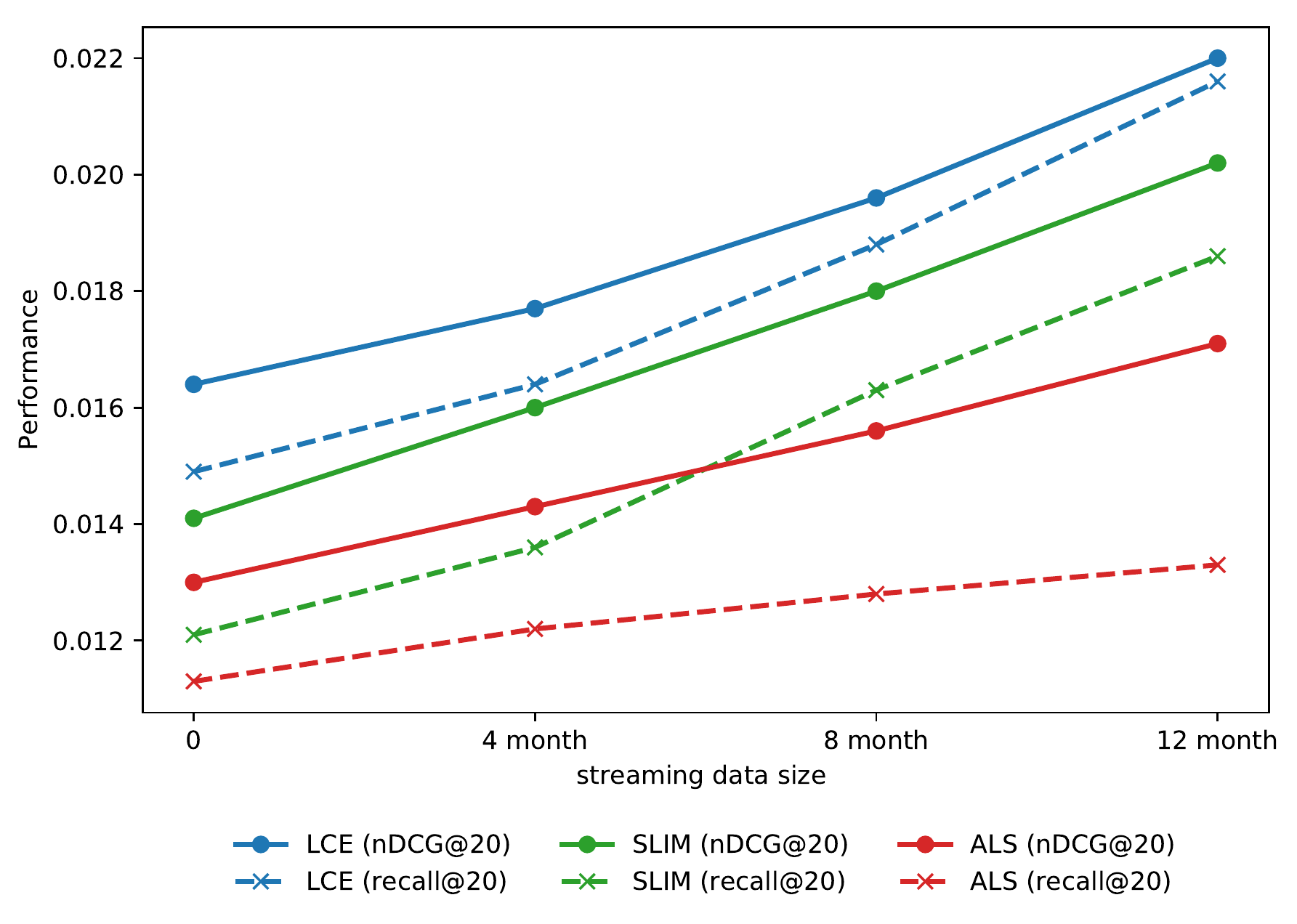}
    \label{fig:cold_start_lthing}
} \\
\subfloat[Epinions]{
    \includegraphics[width = 5.5cm]{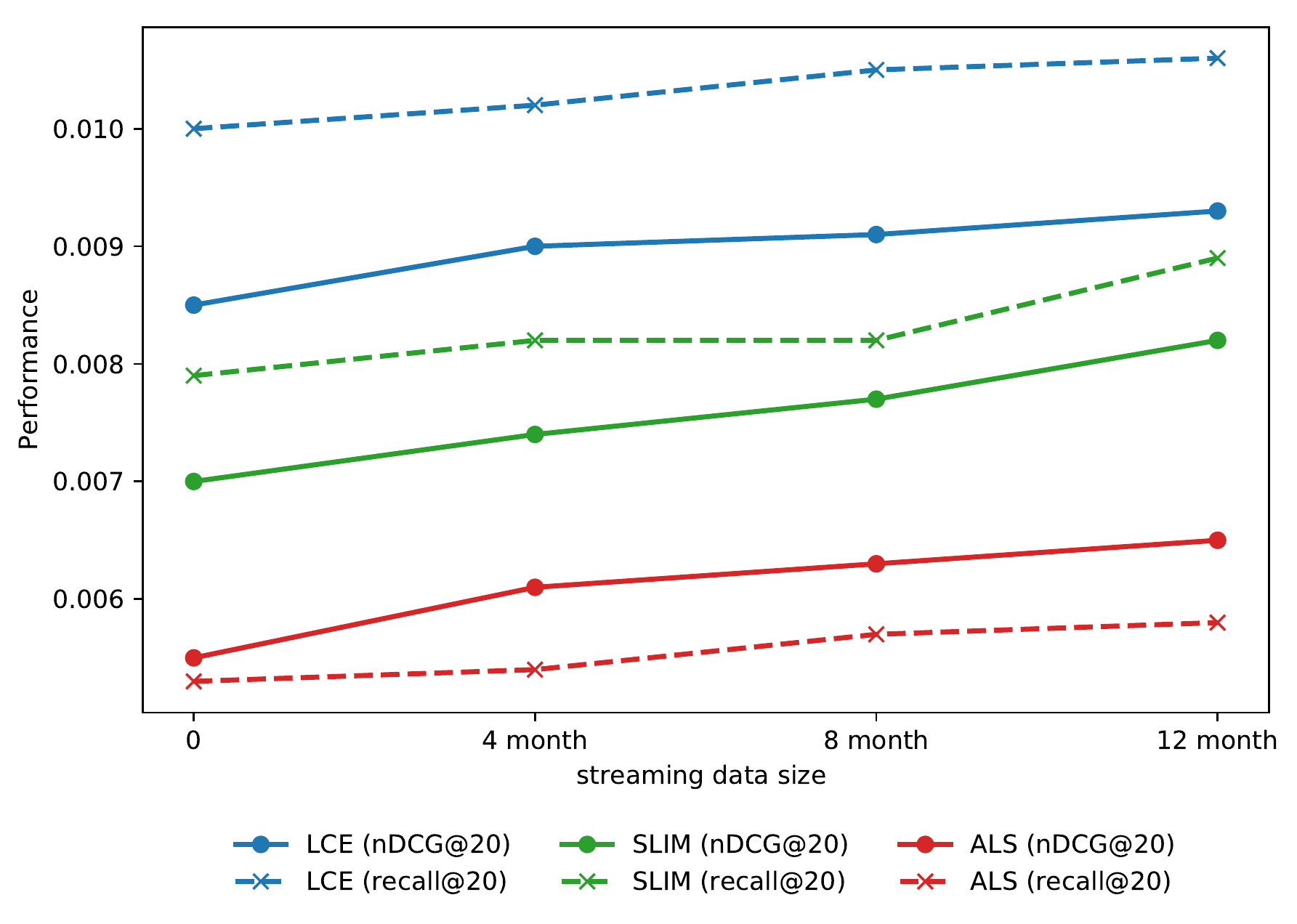}
    \label{fig:cold_start_epin}
}
\vspace{-3.mm}
\caption[]{Recommendation performance considering cold-start items of various methods on LibraryThing (\ref{fig:cold_start_lthing}) and Epinions (\ref{fig:cold_start_epin}). LCE consistently improves the performance on both datasets compared with SLIM and ALS. Note that LightGCN and ENSFM do not support new items.}
\vspace{-8.mm}
\label{fig:cold_start}
\end{figure}

Our model LCE is partially inductive --- it can deal with cold-start users or items depending on the choice of composition direction. We evaluated its recommendation performance when considering cold-start items that only appears during the streaming window on Epinions and Library datasets. Both datasets have a long streaming window (12 months), and contains 40862 and 21506 cold-start items respectively. As LightGCN and ENSFM do not support new items (see~\Cref{tab: model_char}), we compare to two other strong baselines SLIM and ALS. \Cref{fig:cold_start} shows that LCE consistently improves the performance on both datasets, and is able to utilize the streaming data.

\vspace{-2mm}
\subsection{Comparison with LCE variants} \label{sec:expts-ablation}
\vspace{-1mm}

The two important features of our model architecture are (1) using two separate user embeddings for generating item embeddings $(\umat)$ and scoring ($\ummat$) respectively, and (2) stacking multiple graph convolutional layers to capture long-range dependencies in the user-item interaction graphs. We performed two ablation studies to investigate the benefits of the two features. 

In \Cref{tab: results_streaming}, we can see that the single embedding variant (denoted as ``LCE-1 emb'') performs slightly worse than the original LCE, which reveals the benefits of separating the two steps (i.e. embedding generation and scoring). Meanwhile, ``LCE-1 emb'' achieves comparable or better performance than LightGCN, especially in streaming setting, which shows the benefit of introducing compositional embeddings and reconstruction-based objective function. 
Furthermore, the single GCN layer variant (denoted as ``LCE-1 layer'') also achieved suboptimal performance compared to the original LCE with three layers, which shows the effectiveness of modeling long-range dependency in the graph by applying multiple GCN layers. 
Note that the number of layers is also a hyper-parameter, and we can fully expect to further boost the LCE performance by searching in a larger space.

Besides, the choice of composition function $f(\cdot)$ is a hyper-parameter in our model. We compared two common composition functions $avg$ and $sum$ on the three datasets, with the performance shown as $LCE_{mean}$ and $LCE_{sum}$ in \Cref{tab: results_streaming}. We can see that both variants achieved comparable performance. 

\vspace{-2mm}
\subsection{Explicit v.s. implicit embeddings}
\label{sec:explitit_vs_implicit}
\vspace{-1mm}
\begin{table}[]
    \centering
    \begin{tabular}{c|c|c|c}\toprule
         & Yelp & LibraryThing & Epinions \\ \hline
        LCE-\textcolor{blue}{item} & 0.0387$^\star$ & 0.0283$^\star$ & 0.0148$^\star$\\
        LCE-\textcolor{blue}{user} (equal $D$) & 0.0320 & 0.0235 & 0.0109\\
        LCE-\textcolor{blue}{user} (larger $D$) & 0.0322 & 0.0248 & 0.0111\\ \bottomrule
    \end{tabular}
    \caption{The recommendation performance of LCE with different implicit (compositional) embeddings listed in blue. LCE-item is the best performing LCE in \Cref{tab: results_streaming}. Numbers with $\star$ represent significant improvement in a paired t-test at the $p < 0.05$ level compared with the best baseline.}
    \label{tab:lce_user_results}
    \vspace{-8mm}
\end{table}

\begin{figure}
    \centering
    \vspace{-3mm}
    \includegraphics[width=6cm]{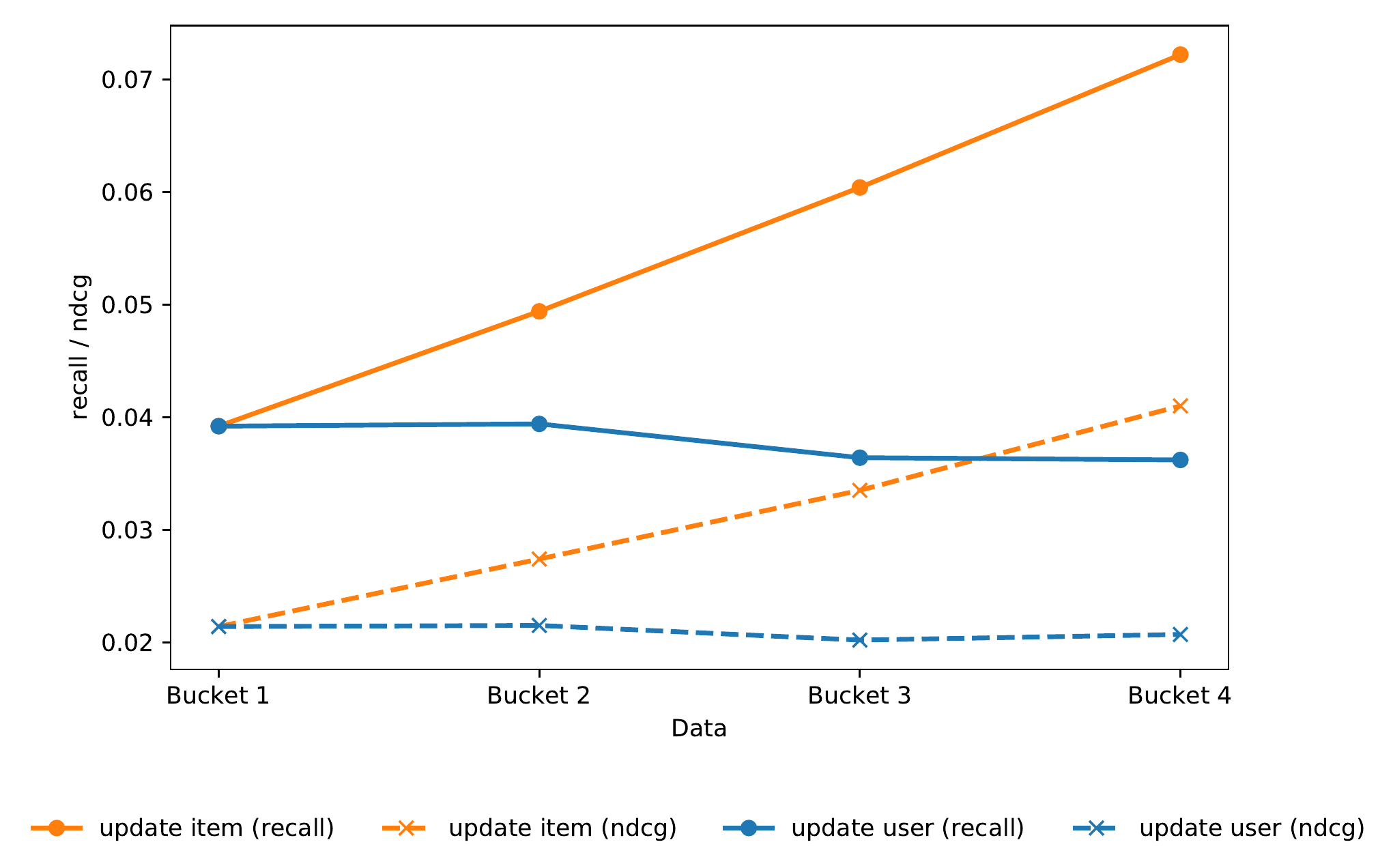}
    \vspace{-3mm}
    \caption{Recommendation performance on Yelp dataset when updating embeddings for only user or item nodes. The superior performance of updating item embeddings validates our assumption that items are more non-stationary.}
    \label{fig:test_stationary}
    \vspace{-3mm}
\end{figure}

One important assumption of our model is that one node type is more amenable to ``static'' representation than the other, so we learn a static representation for that node type (e.g, users) and learn an implicit, {\em compositional} function to calculate the representation of the other nodes (e.g, items). 
In our earlier experiments, we considered the smaller node set to be the more static one, which is the set of users in all the three datasets. We also tried the reverse (i.e. explicitly embedding items and using a compositional function for users). Let the embedding dimension of LCE-item be $d_{item}$. Since the LCE-item model learns two explicit embeddings for each user, the number of parameters is $D=2 d_{item} \: |users|$. When we learn explicit embeddings for items (LCE-user (equal $D$)), we set $d_{user} = \frac{|users|}{|items|}\times d_{item}$ so the total number of parameters is held constant, ie. $2 d_{user} \: |items| = D$. To measure the effect of increasing the parameterization (since the $|items|>|users|$) we also consider $d'_{user}=2d_{user}$ for LCE-user (larger $D$).
The results are shown in \Cref{tab:lce_user_results}, where we can see implicitly embedding users (``LCE-user'') performed significantly worse, regardless of whether the parameter size is increased. 



In addition, we test the assumption of stationarity by measuring the robustness of past embeddings. Since measuring stationarity directly  is challenging, we use robustness as a proxy. Specifically, we split the training data into four equal-sized buckets $\{D_1, D_2, D_3, D_4\}$ in chronological order, and then we use $D_1$ to learn explicit users embeddings $Z^U_1$ and item embeddings $Z^I_1$ with a LightGCN model. To test  robustness, we fix the embedding for one node type (i.e. user embedding $Z^U_1$) and re-learn the embeddings for the other type (i.e. item embedding $Z^I_2, Z^I_3, Z^I_4$) on data $D_2, D_3$ and $D_4$ respectively. For evaluation, the learned embeddings are used to predict the future edges between users and items.

In \Cref{fig:test_stationary} we can see that updating item embeddings (i.e. fix user embeddings) achieved consistently better performance than updating user embeddings, which means the information loss from using old data is more significant for items, and thus we conclude that item nodes are more non-stationary in these datasets. This provides empirical validation of our assumption that an implicit, {\em compositional} function can be used to represent node types that are more non-stationary (to update their representation dynamically) and that it is only necessary to learn explicit representations for node types that are more stationary.

\vspace{-2mm}
\subsection{Evaluation on real-world production data} \label{sec:expts-prod}

In additional to the three public datasets, we also evaluated LCE on a real-world production system (Microsoft Teams), where the task is to recommend TEAMs for users to join. We created a subset of 224K users and 26K TEAMs, with a total of 65 million user interactions. The interaction graph is constructed from a set of time-stamped user-item and user-user actions, where the edges represent interactions between two users (accessing a shared document, email communication, chatting), or between a user and a TEAM (past membership). The offline, streaming and test time windows are set to 3 weeks, 2 weeks and 5 months respectively, and we use daily slices for streaming data. 

For model comparison, we chose LightGCN as the strongest baseline, and evaluated the offline and streaming performance in terms of precision@3 and recall@10. The results showed that LCE achieved competitive offline performance with $> 80\%$ fewer model parameters than LightGCN, and out-performed LightGCN by up to 6.42\% on precision@3 when streaming data is available. Furthermore, our results showed that LCE improves with more streaming data, whereas LightGCN was not able generalize to new data, possibly due to over-parameterization.

%% file: 5_related.tex
\vspace{-2mm}
\section{Related Work}

Our work is inspired by and contributes to three research directions: sequential and streaming recommendation, graph-based recommendation, and cold-start recommendation. 

\vspace{-1mm}
\subsection{Sequential and streaming recommendation}
\vspace{-1mm}

Sequential recommendation models~\cite{rendle2010factorizing, chen2018sequential, tang2018personalized, ma2019hierarchical, xu2019recurrent, qiu2020gag, xu2020graphsail} focus on capturing the temporal dynamics of user-item interactions which is orthogonal to the problem of incremental recommendation. Moreover, these methods cannot handle non-bipartite graph data which we study in this paper. 
Regarding streaming recommendation, prior work~\cite{chang2017streaming, wang2018streaming, diaz2012real} has proposed retraining models with subsamples of the training set~\cite{wang2018streaming} or meta-learning~\cite{zhang2020retrain}. However, so far, these techniques are only shown to be effective for shallow MF models that do not capture higher-order user-item relationships. Due to the large parameter space and high computational cost of deep models or graph-based models, retraining-based methods are not directly applicable. Our work advances the state-of-the-art by developing a graph-based recommendation model that supports incremental recommendation. 

\vspace{-1mm}
\subsection{Graph-based recommendation models} 
\vspace{-1mm}
Leveraging graphs to support recommendation scenarios is an emerging research theme. The core idea is to cast recommendation problem to graph link prediction. Prior work explored many graph neural network architectures, for example GraphSage~\cite{hamilton2017inductive, ying2018graph}. NGCF~\cite{wang2019neural} and LightGCN \cite{he2020lightgcn} learn convolutional structures over embeddings for all users and items. These models are not generally applicable to the incremental streaming recommendation setting that this paper studied, because they require either auxiliary features~\cite{hamilton2017inductive, ying2018graph} or explicit embeddings for all nodes~\cite{wang2019neural, he2020lightgcn}. 
Our work follows the line of graph-based recommenders, but the novel design of compositional embedding allows LCE to readily incorporate new data incrementally. Our experiments show that LCE significantly outperforms state-of-the-art baselines like LightGCN~\cite{he2020lightgcn}. 

\vspace{-1mm}
\subsection{Cold-start recommendation} 
\vspace{-1mm}
Cold-start is a common problem faced by recommendation systems~\cite{maltz1995pointing,schein2002methods}, and it refers to the difficulty of recommending items for users with a limited number of interactions. 
Similar to the compositional embedding proposed in this paper, some prior non-graph models~\cite{liang2018variational, sedhain2015autorec, shi2020beyond} represented users using the items they interacted with to address the cold-start problem. However, the cold-start problem is complementary to the streaming setting that we study in this paper, as it assumes {\em static} setting where recommendation performance is plotted against the number of observations that a user or item has.
Our proposed method LCE can deal with new entities (users or items) as long as they are in the set of nodes whose embeddings are generated compositionally, and we show that LCE is able to improve over the performance of MF and SLIM in the presence of cold-start items. Moving towards a fully inductive setting where both new users and items are considered, the compositional function can be defined on a set of stationary objects or node attributes.

 

%% file: 6_conclusion.tex
\vspace{-1mm}
\section{Conclusion}
Motivated by needs of real-world recommender systems, we consider recommendation in an \emph{incremental streaming setting} in this work, where new data continuously comes in after the model is trained. This is in contrast to the static setting that most graph-based recommendation systems adopt where all information about test nodes are available at training time. We propose a compositional graph-based method (LCE) that supports continuous updates in a streaming setting under low computational cost. 
To evaluate the proposed method, we conduct experiments on three real-world datasets and demonstrate the superior performance of LCE compared to a set of competitive baselines. In particular, the experimental results show that LCE is able to more effectively utilize streaming data, in many cases approaching skyline performance as if it had been relearned with all available data. Moreover, LCE significantly improves over the other baselines when considering cold-start items. We also found positive results for LCE when applying it in a production recommendation setting in Microsoft Teams. 

%% file: 7_appendix.tex
\newpage
\section{Appendix}

\begin{table*}[tpb]
\resizebox{0.8\textwidth}{!}{
\begin{tabular}{lccccc} \toprule
& \multicolumn{2}{c}{prediction} &\multicolumn{3}{c}{model}\\
\cmidrule(lr){2-3} \cmidrule(lr){4-6}
 & \begin{tabular}{c} supports\\  new items \end{tabular} & 
\begin{tabular}{c} efficient \\incremental updates \end{tabular}
& 
\begin{tabular}{c} top-k \end{tabular} & 
\begin{tabular}{c} supports\\ user-user links \end{tabular} & 
\begin{tabular}{c} works w/o \\ features \end{tabular}
\\ \cmidrule(lr){1-1} \cmidrule(lr){2-2}\cmidrule(lr){3-3} \cmidrule(lr){4-4}\cmidrule(lr){5-5} \cmidrule(lr){6-6} 
GAG \cite{qiu2020gag} & & \checkmark  & & \checkmark  & \checkmark\\
JODIE \shortcite{kang2018self, wang2020next, kumar2018learning} & & & & & \checkmark \\
DySAT \cite{sankar2018dynamic} & \checkmark &  &  & & \checkmark \\
GraphSAIL \cite{xu2020graphsail}& \checkmark & & \checkmark & & \checkmark \\
PinSAGE \cite{ying2018graph} &  & \checkmark & \checkmark &  &  \\
LightGCN \cite{he2020lightgcn} &  & \checkmark?& \checkmark& \checkmark& \checkmark \\
\midrule
MF \shortcite{hu2008collaborative, takacs2011applications}& \checkmark & \checkmark?& \checkmark& \checkmark? & \checkmark \\
SLIM \cite{ning2011slim} & \checkmark? & \checkmark?& \checkmark& \checkmark? & \checkmark \\
ENSFM \cite{chen2020efficient}&  &  & \checkmark & \checkmark?& \checkmark \\
\midrule
LCE & \checkmark &  \checkmark&  \checkmark& \checkmark & \checkmark\\ 
\bottomrule
\end{tabular}
}
\caption{Characterization of related recommendation models. $\checkmark?$ means the model can be modified or applied with minor changes. }
\label{tab: model_char}
\end{table*}

\begin{table*}[ht]
\centering
\resizebox{0.8\textwidth}{!}{
\begin{tabular}{lllllll} \toprule
\cmidrule(lr){2-5}
& \multicolumn{2}{c}{Yelp} &\multicolumn{2}{c}{LibraryThing}
&\multicolumn{2}{c}{Epinions}\\
\cmidrule(lr){2-3} \cmidrule(lr){4-5} \cmidrule(lr){6-7}
&  Recall@20 & nDCG@20 & Recall@20 & nDCG@20 & Recall@20 & nDCG@20 \\ \cmidrule(lr){1-1} \cmidrule(lr){2-2}\cmidrule(lr){3-3} \cmidrule(lr){4-4}\cmidrule(lr){5-5}
\cmidrule(lr){6-6}\cmidrule(lr){7-7}
Top-popu & 0.0040 & 0.0023  & 0.0007 & 0.0012 & 0.0011 & 0.0007\\
$RP^3_{\beta}$ & 0.0253 & 0.0128 & 0.0039 & 0.0033 & 0.0046 & 0.0027\\
ALS & 0.0436 & 0.0259 & 0.0174 & 0.0175 & 0.011 & 0.0093\\
SLIM & 0.0481 & 0.0286 & 0.0214 & 0.0198 & 0.0138 & 0.0123\\
ENSFM   &0.0503 & 0.0302 & 0.0199 & 0.0185 & 0.0095 & 0.0068\\
LightGCN    & \textbf{0.0572} & 0.0336 & 0.0237 & 0.0204 & 0.0152 & 0.0096\\
\midrule
$LCE_{mean}$ & 0.0568 & \textbf{0.0338} & \textbf{0.0246} & 0.0213 & 0.0163 & 0.0136\\
$LCE_{sum}$ & 0.0542 & 0.0314 & 0.0244 & \textbf{0.0218}$^\star$ & \textbf{0.0170}$^\star$ & \textbf{0.0144} \\
LCE-1 emb   & 0.0563 & 0.0336 & 0.023 & 0.0206 & 0.0152 & 0.0096\\
LCE-1 layer & 0.0550 & 0.0319 & 0.0186 & 0.0173 & 0.009 & 0.0066\\ \midrule
\% improvement & -0.69\% & +0.60\% & +3.79\% & +6.86\% & +11.84\% & +17.07\%\\
 \bottomrule
\end{tabular}
}
\caption{The offline performance on three datasets. Our model (LCE) consistently out-performs all the other methods. LightGCN and SLIM are the two strongest baselines. Numbers with $\star$ represent significant improvement in a paired t-test at the $p < 0.05$ level compared with the best baseline.}
\label{tab: results_offline}
\end{table*}

\subsection{Comparison with related methods} \label{sec: model_comp}
While there are some existing top-k recommendation methods that can be applied in an incremental streaming scenario, they have clear limitations. Many common methods, e.g., matrix factorization-based models~\cite{hu2008collaborative, mnih2007probabilistic} have no concept of user-user edges and thus have to use heuristics to accommodate such information. Another limitation is that many methods are transductive by nature and do not support new items/nodes at test time. Incremental inference can be added on as a posthoc modification, e.g., through \emph{folding in} \cite{berry1995using}, but this can be ineffective due to the violation of modeling assumptions.

We summarize the properties of several graph-based and traditional recommender systems along these dimensions in \Cref{tab: model_char}. The table shows that our method (LCE) is the only method designed for top-k recommendation in streaming graph scenarios with user-user interaction links, which natively supports predictions over new items and efficient incremental updates, and doesn't require node/item features.

\subsection{Training procedure}
We use the default Xavier initializer \cite{glorot2010understanding} to initialize the model parameters $(\umat, \ummat)$, and train our models using Adam optimizer with weight decay. The model is trained for a maximum of 800 epochs with early stopping, i.e., stopping training if recall@20 on the validation data (the last $10\%$ of offline data) 
does not increase for 50 successive epochs. We also search the best hyperparameters using validation set. Specifically, the embedding dimension is searched in $\{16, 32, 64, 128, 256, 512\}$, no. of GCN layers $L$ is set to $3$, batch size is tuned in $\{2048, 5000, 10000\}$, and the weight decay factor $\lambda$ is tuned in $\{10^{-3}, 10^{-4}, 10^{-5}\}$.

\subsection{Datasets}
\begin{description}
\item[Yelp.] We apply the same 10-core setting as previous papers~\cite{he2020lightgcn, wang2019neural} , i.e., retaining users and items with at least ten interactions.

\item[Epinions.] We select the products from top 10 categories from the year of 2002 to 2011, and filter out users that were inactive since 2004.

\item[LibraryThing.] We filter out users that were inactive since 2007, and only keep users/items with at least 2 reviews before the end of 2009. 
\end{description}

\subsection{Baselines}
\begin{itemize}[leftmargin=*]
    \item Top-Popu: always recommend most popular items excluding already interacted ones.
    \item Alternative Least Squares (ALS) \cite{hu2008collaborative, takacs2011applications}: We applied this common matrix factorization based method to a user-(item $\cup$ user) matrix that was created by concatenating the user-user interaction matrix with the user-item matrix. We first train a model using the offline data, and use the learned matrix as a basis for streaming evaluation. New edges are being incorporated via fold-in. The hyper-parameter embedding dimension $d$ is searched in $\{32, 64, 128, 256, 512\}$. 
    \item Efficient Non-sampling Factorization Machines (ENSFM) \cite{chen2020efficient}. We represent the user-item graph as one-hot feature vectors and increment the counts during online incremental updating. The hyper-parameter negative weight is tuned in $\{0.1, 0.2, 0.5\}$. 
    \item $RP^3_{\beta}$ \cite{christoffel2015blockbusters}: a graph vertex ranking recommendation method that re-ranks items based on 3-hop random walk transition probabilities. The hyper-parameter neighbor size $k$ is searched in $\{50, 100, 200, 500\}$. 
    To further include the user-user interactions in the model when the social network is available, we use a 4-hop random walk with type user-user-item-user instead of the original 3-hop path user-item-user. 
    \item Sparse Linear Method (SLIM) \cite{ning2011slim}. SLIM is one of the most competitive baselines in top-$N$ recommendation. In our experiments, the binary user - (item $\cup$ user) matrix is used as input to the model, and we perform incremental inference by using the learned item-item matrix $A$ with the updated adjacency matrix. The l1 and l2 coefficients are tuned by grid search in $l1 = \{0.01, 0.1, 0.5, 1, 2, 5, 10\}$, $l2 = \{0.1, 0.5, 1, 2, 5, 10, 20\}$.
    \item LightGCN \cite{he2020lightgcn}: LightGCN is a state-of-the-art graph-based model for collaborative filtering. To leverage the social network information, we feed the heterogeneous graph including both user-item and user-user edges instead of the bipartite graph considered in the original paper. For incremental updates, the new updated graphs are fed to the model. The embedding dimension is tuned in $\{16, 32, 64, 128, 256, 512\}$, \# layers = 3, batch size tuned in $\{2048, 5000, 10000\}$, and the weight decay factor is tuned in $\{1e^{-3}, 1e^{-4}, 1e^{-5}\}$. 
    \item LCE variants: In our ablation study (see \Cref{sec:expts-ablation}), we compare to two variants of our model: "LCE-1 emb" and "LCE-1 layer". "LCE-1 emb" does not have a separate user embedding for scoring user-item pairs, but uses the user embedding generated by GCN layers instead (i.e., setting $\uvec$ with \Cref{eq: item_avg}). 
    "LCE-1 layer" uses a single GCN layer instead of three layers. 
\end{itemize}

\begin{figure}
  \centering
  \includegraphics[width=6.5cm]{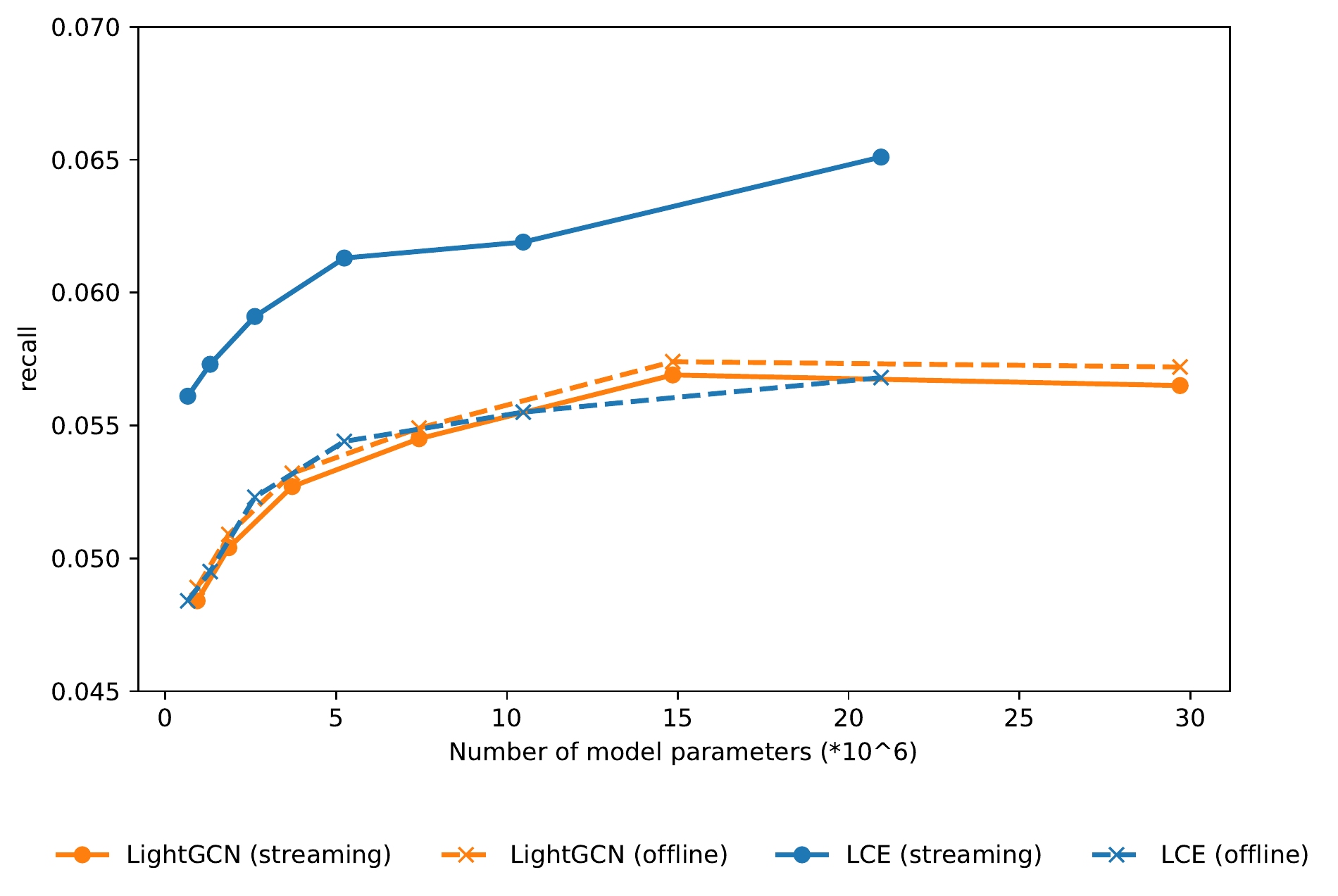}
  \captionof{figure}{Recommendation performance of LCE and LightGCN varying model size (i.e. embedding dimension) on Yelp. LCE refers to $LCE_{mean}$ which has better validation performance than $LCE_{sum}$ on Yelp. LCE achieves competitive offline performance (see dotted lines) with a much smaller model size, and clearly outperforms LightGCN when streaming data is available (see solid lines).}
  \label{fig:model_size_yelp}
\end{figure}

\subsection{Model capacity v.s. performance} \label{sec: capacity}

Note that under the same embedding dimension $d$, LCE has much fewer model parameters than LightGCN. As shown in \Cref{tab:method_char}, LCE learns $2\times |U| \times d$ parameters, while LightGCN learns $(|U| +  |W|)\times d$ parameters where $|U|$ and $|W|$ are numbers of users and items. For example, on Epinions dataset where we have 10,277 users and 87,791 items, LightGCN has about five times more parameters than LCE. \Cref{fig:model_size_yelp} shows the recommendation performance of LCE and LightGCN on the Yelp dataset while varying embedding dimension. We can see that LCE achieves comparable or slightly better offline performance than LightGCN (see dotted lines), and the gain is much larger when streaming data is available (see solid lines), with a much smaller model size.